\documentclass[11pt]{article}
\pdfoutput=1

\usepackage[T1]{fontenc}
\usepackage[utf8]{inputenc}
\usepackage{times}
\usepackage{latexsym}
\usepackage{inconsolata}
\usepackage[final]{ACL2023} 
\usepackage{ACL2023} 
\usepackage{enumitem}

\usepackage{microtype}

\usepackage{amsmath}
\usepackage{amssymb}
\usepackage{amsthm}

\usepackage{algorithm}
\usepackage{algpseudocode}

\usepackage{booktabs}
\usepackage{multirow}
\usepackage{graphicx} 
\usepackage{graphicx}
\usepackage{tikz}
\usepackage{tabularx} 

\usepackage{xcolor}

\usepackage{hyperref}

\usepackage{natbib}

\usepackage{xcolor}
\usepackage{array} 
\usepackage{tabularx} 
\usepackage{multirow}  
\usepackage{float}  
\usepackage{fancyvrb}
\usepackage{tabularx}  
\usepackage{threeparttable}
\usepackage{makecell}
\usepackage{booktabs}
\usepackage{adjustbox}
\usepackage{tcolorbox}

\usepackage{amsmath}

\usepackage{listings}
\usepackage{xcolor}

\title{Adaptive Cost-Efficient Evaluation for Reliable Patent Claim Generation}

\author{
Yongmin Yoo, \quad Qiongkai Xu, \quad Longbing Cao \\
Frontier AI Research Centre, Macquarie University \\
School of Computing, FSE, Macquarie University\\
\texttt{yooyongmin91@gmail.com} \quad 
\texttt{\{qiongkai.xu, longbing.cao\}@mq.edu.au}
}

\begin{document}
\maketitle
\begin{abstract}
Automated patent claim validation demands low error tolerance. However, existing approaches face a rigidity-resource dilemma: lightweight encoders cannot track long-range legal dependencies, while exhaustive LLM verification incurs 4-5$\times$ higher overhead at million-claim scale. A naive confidence-based cascade cannot resolve this because binary validity scores fail to distinguish structurally distinct error types which require different reasoning depths. We propose a two-stage framework: Adaptive Cost-efficient Evaluation (ACE), which exploits the categorical structure of patent errors for uncertainty-aware routing. In the first stage, a fine-tuned encoder projects claims into a $K$+1 distribution over legal error types, whose predictive entropy serves as the routing signal. Claims exceeding an entropy threshold are escalated to the second stage, where an expert LLM executes a schema-constrained Chain-of-Patent-Thought (CoPT) protocol to map claim elements against 35 U.S.C. standards whose schema constraint reduces per-claim latency by 42\% while producing legally grounded verdicts. We further present a 40,000-claim dataset ACE-40k with MPEP-grounded annotations, where ACE surpasses competitive baselines including a supervised 70B-parameter LLM while reducing costs by 78\%. On real USPTO rejection data, the routing mechanism transfers without re-calibration, reducing inference time by 60\% while maintaining competitive recall.
\end{abstract}

\section{Introduction}

Validating generated patent claims poses a uniquely high-stakes challenge in natural language processing. Unlike general prose, patent claims simultaneously function as technical descriptions and legally binding definitions of intellectual property rights. In this domain, a single word choice error, antecedent inconsistency, or logical contradiction can render a claim legally defective, with significant financial and legal repercussions~\citep{mpep2173,faber2023}. Consequently, automating the validation of generated patent claims requires rigorous adherence to legal and structural standards, such as antecedent consistency and logical soundness, while far exceeding the approaches for evaluating standard text generation.

\begin{figure}[t]
  \centering
  \includegraphics[width=0.99\linewidth]{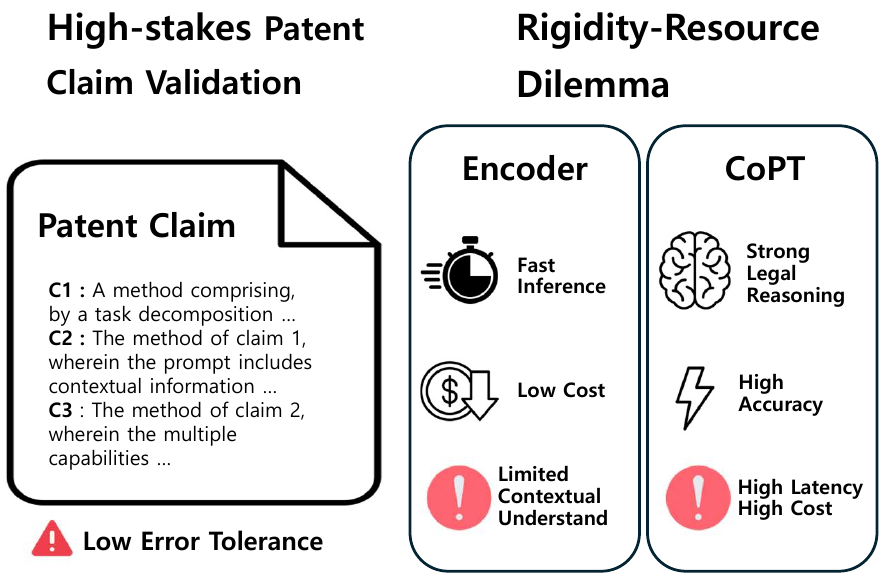}
  \caption{High-stakes patent claim validation requires low error tolerance while demanding cost efficiency. Adaptive Cost-efficient Evaluation (ACE) leverages the throughput of encoder models and the inference accuracy of Chain-of-Patent-Thought (CoPT) reasoning.}
  \label{fig:intro_overview}
  \vspace{-4mm}
\end{figure}

With these stringent requirements, existing text generation evaluation methodologies fail to reconcile validation accuracy with computational scalability. Reference-based metrics, including BLEU~\citep{papineni2002bleu}, ROUGE~\citep{lin2004rouge}, and BERTScore~\citep{zhang2019bertscore}, rely on surface-level n-gram overlap or semantic embedding similarity. These metrics are not designed to diagnose structural flaws, such as antecedent basis violations or circular dependencies, which are critical for legal validity~\citep{mpep2173}. Conversely, while the LLM-as-a-Judge approach offers stronger reasoning capabilities, utilizing LLMs as unconstrained evaluators introduces a critical conflict between industrial-scale throughput and the high reliability required in legal review. First, the extreme computational demand of large-scale models creates a significant scalability bottleneck for exhaustive verification of millions of patent claims. Second, even when resources are available, the inherent stochasticity of generative systems without structured legal grounding often produces plausible but incorrect justifications. Consequently, this combination of substantial computational overhead and probabilistic instability falls short of the reliability and transparency requirements of legal review.

To mitigate the rigidity-resource dilemma, checklist-based evaluations offer a transparent alternative by decomposing complex legal validity criteria into granular verification items, thereby ensuring the interpretability needed in legal review~\citep{ribeiro2020checklist, lee2024checkeval}. However, current implementations typically enforce a static level of scrutiny across all instances, regardless of their difficulty. This introduces a redundant computational burden, as unambiguous claims consume the same resources as complex borderline cases. Consequently, an adaptive mechanism is needed that intelligently modulates the depth of reasoning, reserving exhaustive checklist verification for cases exhibiting high intrinsic uncertainty.

To bridge this gap, we propose the \textbf{A}daptive \textbf{C}ost-efficient (patent) \textbf{E}valuation (ACE) framework. The key insight is that the predictive entropy of a lightweight encoder can effectively identify which claims require expensive LLM verification. ACE engineers a fine-tuned PatentBERT-based encoder as an uncertainty-aware gatekeeper whose predictive entropy serves as a runtime routing signal, escalating only high-uncertainty claims to an expert LLM instruction-tuned for legal reasoning. The expert then executes a Chain of Patent Thought (CoPT) protocol, performing structured statutory mapping to produce legally grounded verdicts. By combining efficient screening with selective deep verification, ACE decouples reasoning depth from evaluation cost.

Our contributions are summarized as follows:
\begin{itemize}
    \item \textbf{System-Level Innovation:}  ACE as an entropy-driven hybrid framework decouples reasoning depth from evaluation cost, surpassing competitive baselines including a 70B-parameter LLM while reducing operational costs and inference time by 78\%.

    \item \textbf{Diagnostic Insight:} We identify limitations of encoder-based models in tracking long-range dependencies (e.g., \textit{Antecedent Basis}), then ACE bridges this gap via selective generative auditing. Analysis on real prosecution data further reveals that cross-claim multi-hop references constitute the dominant residual failure mode.

    \item \textbf{Reasoning Protocol:} We introduce the Chain of Patent Thought (CoPT), a schema-constrained protocol that structures reasoning around 35 U.S.C. standards, reducing inference latency by 42\% (11.85s $\rightarrow$ 6.88s) while producing actionable diagnostic outputs for legal practitioners.

    \item \textbf{Domain Datasets:} We introduce ACE-40k, a 40,000-claim dataset with hierarchical labels grounded in MPEP error categories for training and primary evaluation. We additionally present ACE-Real112b, a corpus of 204 genuine USPTO §112(b) rejections that serves as an out-of-domain stress test, on which ACE's routing mechanism transfers without re-calibration.
\end{itemize}

\section{Related Work}

\subsection{General Evaluation and Scalability Barriers}
\label{sec:related_general}

Standard metrics, ranging from lexical n-gram overlaps~\citep{papineni2002bleu, lin2004rouge, banerjee2005meteor} to embedding-based measures~\citep{zhang2019bertscore, zhao2019moverscore, sellam2020bleurt}, exhibit a fundamental semantic-legal gap in patent contexts. Patent validity hinges on rigid structural dependencies; high semantic overlap routinely masks fatal legal flaws such as antecedent basis violations or term-of-art misuse~\citep{mpep}. The LLM-as-a-Judge approach~\citep{zheng2023judging, liu2023geval, fu2024gptscore} and checklist-driven decompositions~\citep{min2023factscore, lee2024checkeval} offer stronger diagnostic capacity, yet apply uniform computational depth to every instance, rendering exhaustive validation of millions of claims economically unsustainable.

Adaptive computation has been explored through early-exit decoding~\citep{schuster2022calm}, LLM cascading~\citep{chen2024frugalgpt}, and selective prediction that abstains or defers based on confidence signals~\citep{varshney2022selective}. However, these methods route on binary or scalar confidence, which is poorly calibrated when the underlying task involves multiple heterogeneous failure modes of differing legal severity. ACE derives its routing signal from a K+1 categorical entropy, enabling escalation decisions that reflect error-type heterogeneity rather than undifferentiated uncertainty.

\subsection{Domain-Specific Constraints}
\label{sec:related_patent}

Domain-adapted encoders such as PatentBERT~\citep{lee2020patentbert} and evaluation frameworks including PatentScore~\citep{yoo2025patentscore} and PEDANTIC~\citep{knappich2025pedantic} address patent-specific syntax and legal conventions. However, PatentScore condenses structural flaws into scalar scores without actionable diagnostic rationale, while PEDANTIC restricts its scope to definiteness, leaving broader defects such as circular dependencies unaddressed. Crucially, both process trivial and complex claims with identical depth. ACE integrates an uncertainty-driven gatekeeper with a selective LLM evaluator, reserving exhaustive statutory verification for the minority of claims that genuinely require it.

\section{Dataset Construction}
\label{sec:dataset}

\begin{table}[h]
\centering
\small
\resizebox{0.99\columnwidth}{!}{
\begin{tabular}{llr}
\toprule
\textbf{Category} & \textbf{Description} & \textbf{Count} \\
\midrule
\textbf{Pass} & Published patent & \textbf{20,000} \\
\midrule
Fail: Antecedent & Lack of antecedent basis & 4,000 \\
Fail: Dependency & Invalid cross-reference & 4,000 \\
Fail: Logical & Impossible causality or circularity & 4,000 \\
Fail: Ambiguity & Indefinite terminology & 4,000 \\
Fail: Syntax & Structural format violations & 4,000 \\
\midrule
\textbf{Total} & \textbf{Balanced Hierarchically Labeled Dataset} & \textbf{40,000} \\
\bottomrule
\end{tabular}
}
\caption{Statistics of the constructed ACE dataset. We enforce a strictly balanced distribution of 1:1 between valid and invalid samples.}
\label{tab:dataset}
\end{table}

To facilitate robust automated validation and fine-grained error diagnosis for patent claims, we constructed a large-scale, balanced patent claim dataset with hierarchical labels, comprising authentic patent claims and synthetically corrupted counterparts. The corrupted samples were generated using a rule-driven synthetic error injection method powered by multiple LLMs, a strategy specifically adopted to prevent bias towards a single model's generation pattern. Distinct from prior studies predicated on subjective quality metrics, our dataset aligns directly with the standards of the Manual of Patent Examining Procedure~\citep{mpep}, centering on the binary validity of a claim and the specific rationale for rejection. Detailed statistics and representative examples of the constructed samples are provided in Appendix~\ref{sec:appendix_data}.

\subsection{Data Composition and Error Injection}
The dataset is derived from legally granted patents within the USPTO Bulk Data Storage System~\citep{uspto_bulk}, ensuring that all positive samples are drawn from legally granted patents. To synthesize negative samples, we employed a \textit{rule-driven synthetic negative sampling} approach. For each valid anchor claim, we systematically injected targeted errors corresponding to five distinct rejection categories defined in patent law.

Crucially, to prevent model bias towards the majority class and ensure rigorous evaluation, we strictly balanced the distribution between valid and invalid samples. The final dataset consists of 40,000 samples, split evenly between valid claims labeled as Pass and invalid claims assigned to specific error categories labeled as Fail. This ratio of 1:1 ensures that the performance of the model reflects genuine discriminatory capability rather than a heuristic reliance on class priors. The error taxonomy is rigorously grounded in specific MPEP sections, targeting five key categories: Antecedent Basis, Dependency, Logical, Ambiguity, and Syntax. Detailed definitions for these violations and the corresponding generation prompts are provided in Appendices~\ref{sec:error_taxonomy} and~\ref{sec:appendix_data_generation}.

\subsection{Strategic Alignment and Utility}
Our error taxonomy is strategically aligned with known generative failure modes: Antecedent and Dependency errors serve as proxies for hallucinations~\citep{liu2023lost}, while Logical and Ambiguity errors target causal inconsistency~\citep{zhang2023causal} and vagueness~\citep{tam2023ambiguity}. To support the ACE framework, samples are annotated with hierarchical labels (binary validity and fine-grained categories), a structure whose reliability was verified through a rigorous annotation audit. Specifically, two PhD-level patent domain experts independently evaluated a random sample of 500 instances, evenly split between the two generators (Qwen3-32B and Gemma-3-27B-IT), on both binary validity and 6-way hierarchical error labels. The audit yielded a raw agreement of 99.2\% and a Cohen's $\kappa$ of 0.98 with no significant difference in agreement rate between generators, confirming near-perfect inter-annotator consistency and the absence of detectable model-specific artifacts. This dual-granularity is critical for our architecture: the binary labels train the efficient Gatekeeper to detect general inadmissibility, as described later. Simultaneously, the fine-grained categories facilitate uncertainty quantification via entropy calculation and provide fine-grained supervision for the LLM, as detailed subsequently.
\section{Methodology}
\label{sec:methodology}

\begin{figure*}[t]
\centering
\includegraphics[width=0.99\textwidth,keepaspectratio]{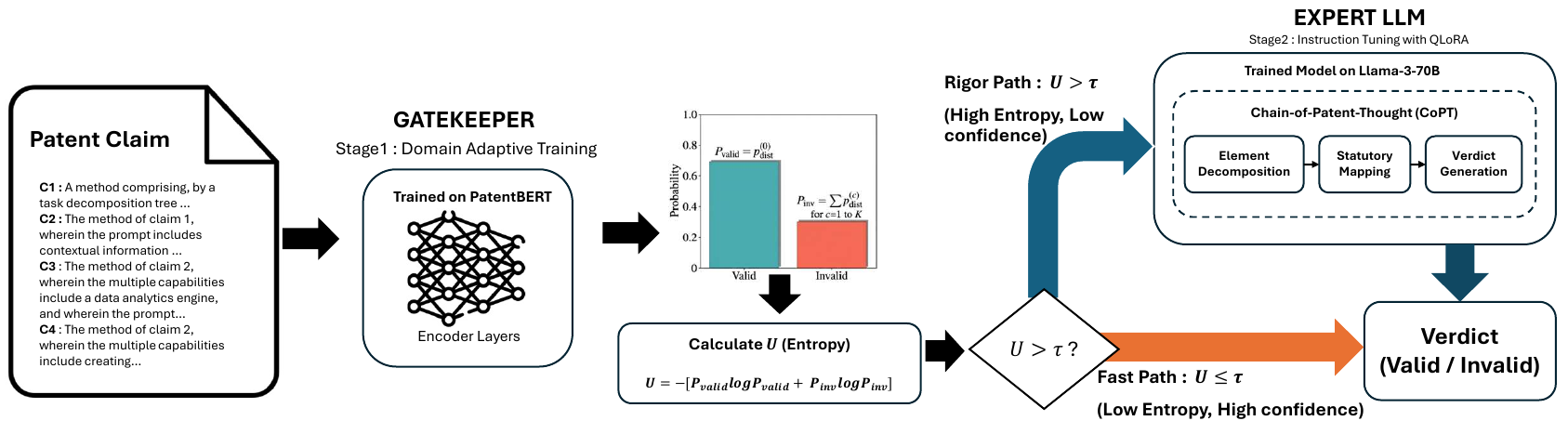}
\caption{The ACE framework. An uncertainty-aware Gatekeeper routes patent claims based on 
predictive entropy: confident predictions ($U \le \tau$) are resolved via the Fast Path, 
while ambiguous claims ($U > \tau$) are escalated to an instruction-tuned Expert LLM that 
performs Chain-of-Patent-Thought (CoPT) reasoning for statutory verification.}
\label{fig:architecture}
\vspace{-4mm}
\end{figure*}

We propose the \textbf{A}daptive \textbf{C}ost-efficient \textbf{E}valuation (ACE) framework, designed to reconcile computational scalability with legal reliability. As illustrated in Figure~\ref{fig:architecture}, ACE integrates a high-throughput PatentBERT gatekeeper with a reasoning-intensive LLM via uncertainty-driven routing. The framework operates in two phases: (1) \textit{domain-adaptive training} to optimize both the uncertainty-aware encoder and the instruction-tuned evaluator, and (2) \textit{adaptive inference}, where high-entropy claims are dynamically escalated to the expert LLM, effectively decoupling evaluation cost from reasoning depth.

\subsection{Domain-Adaptive Training Strategy}
\label{sec:training_strategy}

The ACE framework jointly optimizes two specialized components, ensuring that each model is tailored to its distinct role in the triage pipeline; the specific training hyperparameters for both components are detailed in Appendix~\ref{sec:appendix_training}.

\paragraph{Stage 1: Gatekeeper via Unified Multitask Learning}
We engineer the Gatekeeper using PatentBERT~\citep{lee2020patentbert} to process claims with high throughput. Instead of employing decoupled heads for binary and multiclass tasks, which can lead to feature misalignment, we propose a Unified Categorical Projection. Specifically, we project the latent representation $h$ into a single distribution over $K+1$ classes, encompassing one valid state ($c=0$) and $K$ specific error types ($c=1, \dots, K$),
\begin{equation}
    p_{\text{dist}} = \text{softmax}(W_e h + b_e) \in \mathbb{R}^{K+1}.
\end{equation}
To optimize this granular output space, we employ a standard cross-entropy objective:
\begin{equation}
    \mathcal{L}_{\text{GK}} = -\sum_{c=0}^{K} y_c \log p_{\text{dist}}^{(c)}.
\end{equation}
By training directly on this fine-grained taxonomy, the encoder learns a shared representation where validity is intrinsically modeled as the \textit{absence} of structural defects. This probabilistic granularity is essential for the robust entropy calculation utilized in the subsequent inference stage.

\paragraph{Stage 2: Expert Evaluator via Instruction Tuning}
To resolve complex statutory ambiguities (e.g., 35 U.S.C. §112), we fine-tune Llama-3-70B-Instruct. We adhere to computational constraints (single NVIDIA H100) via Quantized Low-Rank Adaptation (QLoRA). The backbone is frozen in 4-bit NormalFloat (NF4), and low-rank adapters ($r=16, \alpha=32$) are injected into all linear projection layers using \textit{bfloat16} precision. The model is optimized via Supervised Fine-Tuning (SFT) employing a response-only masking strategy, where the training loss is computed exclusively on the generated completion tokens, to prevent catastrophic forgetting of general reasoning capabilities while enhancing domain adherence.

\subsection{Adaptive Inference Framework}
\label{sec:adaptive_inference}

We deploy the trained models within a dynamic routing framework designed to decouple reasoning depth from evaluation cost.

\paragraph{Stage 1: Uncertainty Quantification \& Threshold Calibration}
To assess instance-level difficulty, we first reconstruct the global validity verdict from the Gatekeeper's fine-grained output. Given the distribution $p_{\text{dist}}$, we define the probability of a claim being invalid ($P_{\text{inv}}$) by marginalizing over all specific error subtypes ($c=1 \dots K$):
\begin{equation}
    P_{\text{valid}} = p_{\text{dist}}^{(0)}, \quad P_{\text{inv}} = \sum_{c=1}^{K} p_{\text{dist}}^{(c)}.
\end{equation}
Using these aggregated probabilities, we calculate the predictive entropy $U$ as a runtime proxy for reliability:
\begin{equation}
    U = - \left[ P_{\text{valid}} \log P_{\text{valid}} + P_{\text{inv}} \log P_{\text{inv}} \right].
\end{equation}
This categorical entropy captures error-type heterogeneity that binary confidence scores cannot distinguish. To rigorously determine the optimal routing threshold $\tau$, we utilize a \textit{Risk-Coverage Trade-off Analysis}. We identify the optimal escalation rate $\gamma^*$ by maximizing the trade-off between the Gatekeeper's F1-score $\mathcal{F}(\gamma)$ on retained samples and the escalation ratio $\gamma$,
\begin{equation}
    \gamma^* = \arg\max_{\gamma \in [0,1]} \left( \mathcal{F}(\gamma) - \lambda \cdot \gamma \right),
\end{equation}
where $\lambda$ governs the trade-off preference between accuracy and computational overhead. The threshold $\tau$ is then calibrated to the uncertainty value at the $\gamma^*$-th percentile.

\paragraph{Stage 2: Selective Escalation via CoPT}
The final validity decision $\hat{y}$ is governed by a resource-aware hard-switching policy based on the calibrated threshold $\tau$. For the Fast Path, the binary prediction $p_{\text{bin}}$ is derived as $p_{\text{bin}} = \operatorname{argmax} (P_{\text{valid}}, P_{\text{inv}})$ :
\begin{equation}
    \hat{y} = 
    \begin{cases} 
    p_{\text{bin}} & \text{if } U \le \tau \quad (\text{Fast Path}) \\
    p_{\text{LLM}} & \text{if } U > \tau \quad (\text{Rigorous Path})
    \end{cases}
\end{equation}

Claims located in the high-entropy region ($U > \tau$) are escalated to the Supervised Fine-tuned Expert LLM to execute the Chain of Patent Thought (CoPT) protocol. We reformulate the validity assessment not as a simple classification task, but as a conditional sequential generation problem. The expert model is mandated to first construct a reasoning chain $R$, encompassing element decomposition and statutory mapping against 35 U.S.C. standards, before deducing the verdict $y$. The comprehensive prompt templates and in-context examples utilized for this diagnostic process are detailed in Appendix~\ref{sec:appendix_prompts}. Mathematically, we optimize and select the best reasoning chain and verdict, $\hat{R}$ and $\hat{y}$ by maximizing the joint likelihood:
\begin{equation}
    \arg\max_{R, y} \underbrace{P(y \mid R, x)}_{\text{Verdict}} \cdot \underbrace{P(R \mid x)}_{\text{Reasoning}}.
\end{equation}
By explicitly factorizing the probability, this formulation enforces a strict causal dependency where the final verdict $y$ is conditioned on the diagnostic evidence $R$. In practice, this joint objective is approximated via autoregressive decoding, where the model first generates $R$ token-by-token and then produces $y$ conditioned on the completed reasoning chain. This effectively mitigates ungrounded hallucinations, ensuring that the decision resolves complex structural ambiguities (e.g., Antecedent Basis) that the Gatekeeper's encoder architecture fails to capture. 
\section{Experiments}
\label{sec:experiments} 

We evaluate ACE on the ACE-40k dataset\footnote{The ACE-40k, ACE-Real112b datasets and code will be publicly released upon acceptance.} (8:1:1 split), benchmarking against general encoders (e.g., Longformer), domain-specific encoders (e.g., PatentBERT), and large language models (Llama-3-70B-Instruct). To validate the stability of our uncertainty estimation, we additionally perform stratified 5-fold cross-validation on the Stage 1 Gatekeeper. All models are evaluated on a single NVIDIA H100 80GB GPU. We report Macro-F1 and estimated operational cost (\$3.00/hr); detailed configurations are in Appendix~\ref{sec:appendix_training}.

\begin{table*}[t]
    \centering
    \small
    \renewcommand{\arraystretch}{1.2}
    \setlength{\tabcolsep}{8pt} 
    
    \begin{tabular}{l c c c c c} 
    \toprule
    \textbf{Model / Method} & \textbf{Acc. (\%)} & \textbf{F1 (\%)} & \textbf{AUC (\%)} & \textbf{Latency (s)} & \textbf{Est. Cost (\$)/M} \\
    \midrule
    
    \multicolumn{6}{l}{\textit{Sanity Check \& General Baselines}} \\
    Vanilla BERT                            & 49.60 & 66.25 & 50.93 & 0.12 & \$100 \\
    Vanilla Longformer                      & 75.52 & 73.11 & 47.26 & 0.42 & \$350 \\
    Frozen BERT + Logistic Regression       & 69.27 & 68.12 & 66.20 & 0.08 & \$67 \\
    TF-IDF + Logistic Regression            & 73.38 & 72.23 & 83.00 & $<$ 0.01 & $<$ \$1 \\
    \midrule
    
    \multicolumn{6}{l}{\textit{Domain-Specific Experts}} \\
    Legal-BERT & 69.70 & 68.86 & 46.30 & 0.12 & \$100 \\
    PatentBERT & 77.25 & 76.59 & 51.16 & 0.12 & \$100 \\
    PatentBERT Fine-tuned &91.65 & 91.58 & 97.16 & 0.12 & \$100  \\

    \midrule
    
    \multicolumn{6}{l}{\textit{Large Language Models (Llama-3-70B)}} \\
    Base w/ Zero-shot                   & 70.10 & 67.75 & - & 7.31 & \$6,092 \\
    Base w/ Few-shot                    & 60.00 & 34.85 & - & 3.32 & \$2,767 \\
    SFT w/ Zero-shot                    & 94.60 & 94.63 & - & 11.85 & \$9,875 \\
    SFT w/ CoPT (Expert LLM Only)           & 94.80 & 94.76 & - & 6.88 & \$5,733 \\
    \midrule
    
    \multicolumn{6}{l}{\textit{Proposed Framework}} \\
    \textbf{ACE Gatekeeper (Stage 1 Only)} & 91.65 & 91.58 & 97.16 & 0.12 & \$100 \\
    \textbf{ACE Hybrid (Stage 1 + 2)}      & \textbf{95.05} & \textbf{94.95} & - & \textbf{1.50} & \textbf{\$1,247} \\
    \bottomrule
    \end{tabular}
    
    \caption{Main Performance Results. Domain-specific baselines use pre-trained representations without task-specific fine-tuning. ACE Hybrid processes 1M claims at \$1,247, a 78\% reduction versus the standalone Expert LLM (\$5,733). Cost estimated at \$3.00/hr for NVIDIA H100. AUC is reported only for single-model classifier.}
    \label{tab:main_results}
\end{table*}

\subsection{Main Performance Analysis}
\label{sec:performance_analysis}

Table \ref{tab:main_results} presents the evaluation of the ACE framework against baseline methods. The ACE Hybrid model achieves the best F1 among the evaluated methods at 94.95\%, surpassing competitive baselines while reducing operational costs to \$1,247 per million claims, a 78\% reduction compared to the standalone Expert LLM (\$5,733).

\paragraph{Efficacy and Efficiency of CoPT.}
The standalone expert model (SFT w/ CoPT) not only achieves high F1 (94.76\%) but also demonstrates significantly lower latency (6.88s) compared to the SFT Zero-shot baseline (11.85s). This counter-intuitive efficiency gain stems from our CoPT protocol imposing a strict JSON schema on the output, which curbs uncontrolled verbosity and optimizes both reasoning quality and inference speed.

\paragraph{Breaking the Performance Ceiling.}
Without task-specific fine-tuning, the domain-specific encoder PatentBERT plateaus at 76.59\% F1, confirming that domain-adaptive pre-training alone is insufficient for capturing the long-range dependencies required in patent claim validation. Even after fine-tuning (91.58\% F1), the encoder cannot match the Expert LLM (94.76\%), indicating a fundamental reasoning gap that requires the generative capabilities integrated by ACE.

\paragraph{Discriminative Robustness (AUC).}

\begin{figure}[t]
    \centering
    \includegraphics[width=1.0\linewidth]{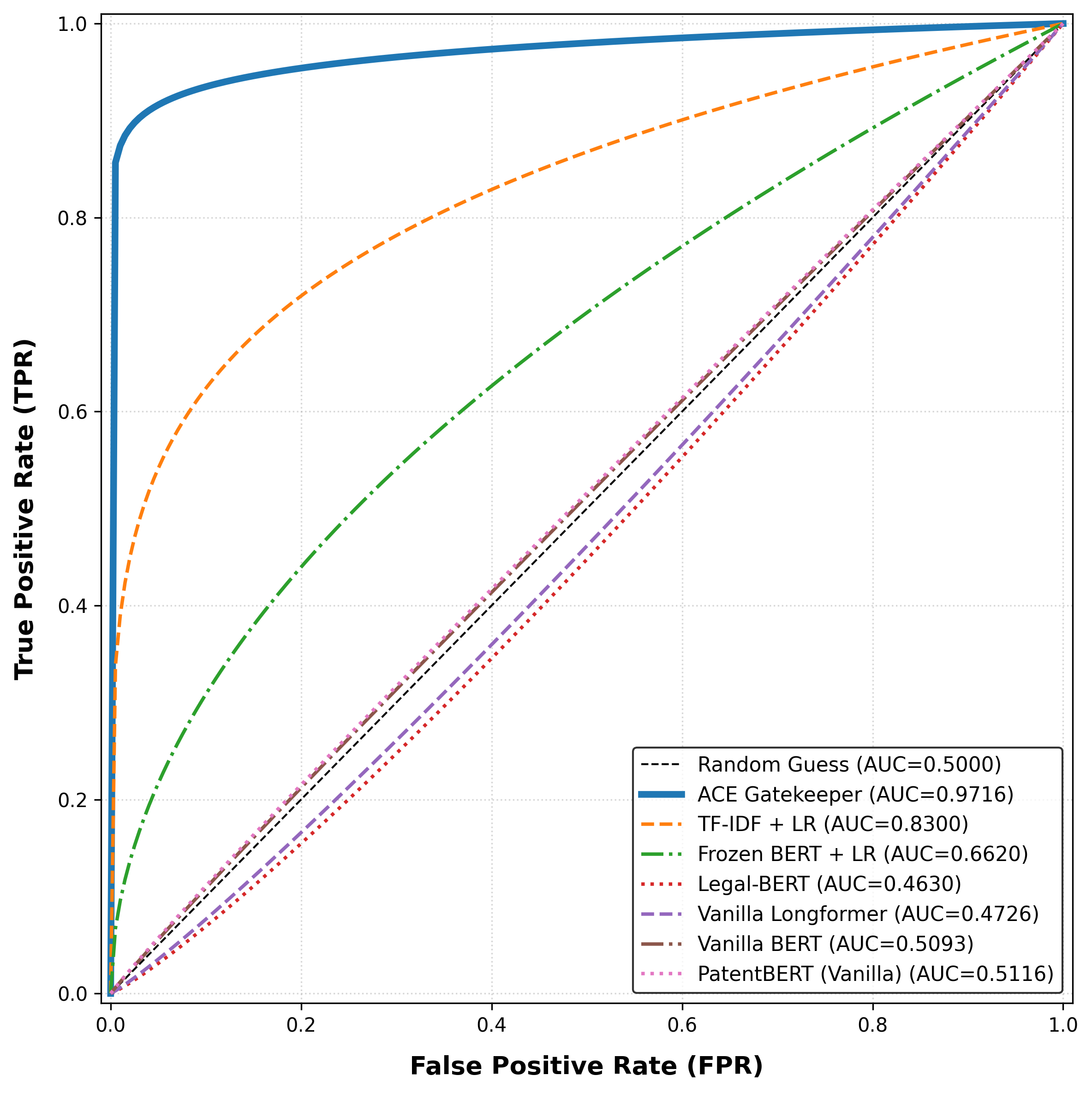}
    \caption{ROC curves for claim validity discrimination. The Gatekeeper (AUC=0.9716) substantially outperforms baselines.}
    \label{fig:roc_curve}
\end{figure}

The Stage 1 Gatekeeper achieves an AUC of 97.16\%, contrasting sharply with vanilla PatentBERT (51.16\%) and Legal-BERT (46.30\%), which perform near random chance. While the lexical baseline (TF-IDF) reaches 83.00\%, it fails to capture the complex logical dependencies that our Gatekeeper masters. This high AUC validates the entropy-based routing logic: the model's confidence scores provide a reliable signal for determining which claims require expensive Stage 2 reasoning.

\subsection{Real-World USPTO Evaluation}
\label{sec:real_world}

While the main results in Section~\ref{sec:performance_analysis} are grounded in the synthetically constructed ACE-40k dataset, validating transferability to authentic patent prosecution is essential. To this end, we introduce ACE-Real112b, a positive-only stress-test comprising 204 genuine \S112(b) rejections from the Harvard USPTO Patent Dataset~\citep{suzgun2023hupd}.

\paragraph{Real World Dataset Construction.}
From the Harvard USPTO Patent Dataset~\citep{suzgun2023hupd}, we extract patent applications filed in 2016 that received at least one \S112(b) indefiniteness rejection in their Office Actions, yielding 204 evaluation instances (one claim per application). Because all samples are authentic rejections (positive-only), we report recall as the transfer metric. No thresholds or parameters were re-tuned on this set; the escalation ratio calibrated on ACE-40k is applied without modification.

\begin{table}[t]
    \centering
    \small
    \renewcommand{\arraystretch}{1.15}
    \setlength{\tabcolsep}{5pt}
    \begin{tabular}{l c c}
    \toprule
    \textbf{Model} & \textbf{Recall (\%)} & \textbf{Time (s)} \\
    \midrule
    Llama-3-8B (zero-shot) & 13.2 & 103.1 \\
    Llama-3-70B (zero-shot) & 8.3 & 355.3 \\
    PatentBERT Standalone & 56.4 & 7.2 \\
    Llama-3-70B + SFT + CoPT & 65.2 & 6{,}820.9 \\
    \midrule
    \textbf{ACE Hybrid (Ours)} & \textbf{66.2} & \textbf{2{,}674.8} \\
    \bottomrule
    \end{tabular}
    \caption{Results on ACE-Real112b (204 real USPTO \S112(b) rejections). All samples are genuine Office Action rejections; recall is reported as the transfer metric. No parameters were re-tuned on this set.}
    \label{tab:real_world}
\end{table}

\paragraph{Results and Analysis.}
As presented in Table~\ref{tab:real_world}, the ACE Hybrid achieves 66.2\% recall, matching the unrouted Expert LLM (65.2\%) while reducing inference time by over 60\%. Compared to PatentBERT standalone, the hybrid framework provides a +9.8 percentage point improvement, confirming that entropy-based routing effectively channels difficult cases to generative reasoning. Notably, zero-shot LLMs (Llama-3-8B: 13.2\%, Llama-3-70B: 8.3\%) fail substantially on real prosecution data, reaffirming that domain-specific fine-tuning and structured prompting are indispensable for \S112(b) detection. The fact that the escalation threshold calibrated on synthetic data transfers directly to real USPTO rejections without re-calibration confirms the distributional robustness of the Gatekeeper's uncertainty estimation. A detailed threshold sensitivity analysis across escalation ratios from 10\% to 50\% is provided in Appendix~\ref{sec:appendix_threshold}, confirming a smooth and monotonic recall-cost trade-off.

\subsection{Fine-grained Error Analysis}
\label{sec:fine_grained}

\begin{figure}[h]
  \centering
  \includegraphics[width=0.95\linewidth]{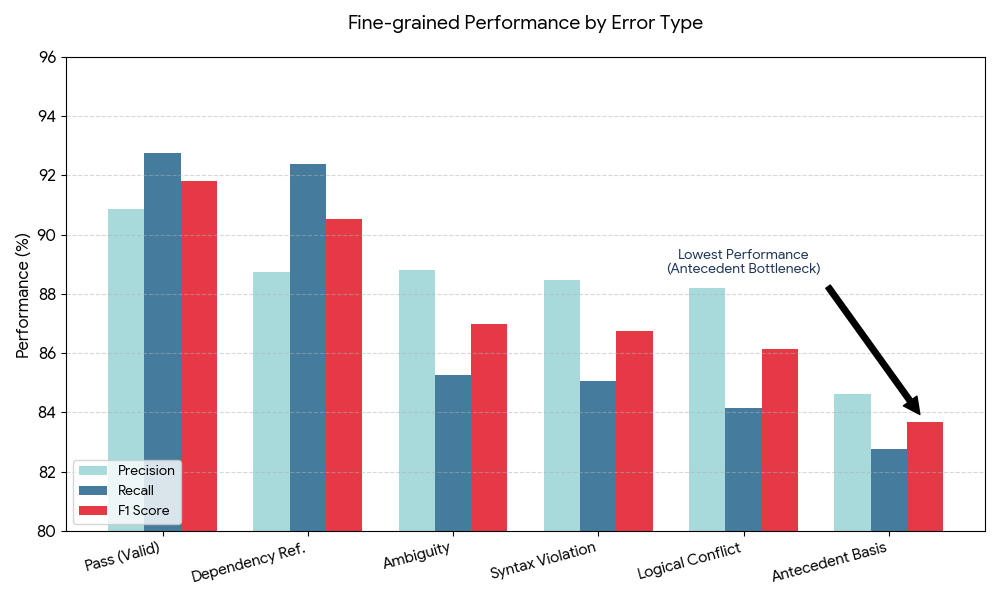}
  \caption{Fine-grained Gatekeeper performance across error types. \textit{Antecedent Basis} records the lowest recall, confirming the long-range tracking bottleneck.}
  \label{fig:error_analysis}
\end{figure}

As shown in Figure~\ref{fig:error_analysis}, the Gatekeeper excels at detecting \textit{Dependency Reference} errors (92.37\% recall), which manifest as explicit pattern mismatches easily captured by self-attention, yet records the lowest recall on \textit{Antecedent Basis} errors (82.77\%), confirming that tracking entity lifecycles (e.g., ``a widget'' $\rightarrow$ ``the widget'') across long sequences exceeds the semantic consistency limits of encoder-based models. This $\sim$17\% miss rate constitutes a significant legal risk that ACE mitigates through selective escalation: the CoPT protocol explicitly identifies such violations via element parsing (e.g., \textit{``Step 2: `the sensor' lacks a proper antecedent basis''}), as illustrated in Appendix~\ref{sec:appendix_copt_example}.

\begin{table}[h]
    \centering
    \small
    \renewcommand{\arraystretch}{1.15}
    \setlength{\tabcolsep}{4pt}
    \begin{tabular}{l c c}
    \toprule
    \textbf{Failure Pattern} & \textbf{Count} & \textbf{\%} \\
    \midrule
    Cross-claim dependency / Nested antecedent & 38 & 55.1 \\
    Claim-set semantic scope ambiguity & 17 & 24.6 \\
    Examiner-specific subjective judgment & 14 & 20.3 \\
    \bottomrule
    \end{tabular}
    \caption{Breakdown of 69 missed cases on ACE-Real112b.}
    \label{tab:real_error}
\end{table}

\paragraph{Real-World Error Patterns.}
Analysis of the 69 missed cases on ACE-Real112b (Table~\ref{tab:real_error}) reveals that the antecedent bottleneck persists as the dominant failure source: 38 errors (55.1\%) stem from cross-claim dependencies where elements introduced in an independent claim propagate through chains of dependent claims, forming multi-hop references spanning the entire claim set. The remaining 31 errors divide into claim-set semantic scope ambiguity (17 cases, 24.6\%), where validity requires access to the full specification beyond claim text, and examiner-specific subjective judgment (14 cases, 20.3\%), where rejection rests on interpretive ambiguity rather than explicit structural defects. The former delineates a principled input-scope boundary, while the latter reflects inherent label indeterminacy, together suggesting that the gap between synthetic (94.95\% F1) and real-world (66.2\% recall) performance stems primarily from a qualitative shift in error complexity, from intra-claim single errors to multi-hop cross-claim dependencies (55.1\% of failures), rather than a generalization failure of the routing mechanism. We further validated ACE's stability through stratified 5-fold cross-validation ($\sigma_{\text{AURC}} < 0.002$) and a simulated 9:1 industrial distribution, which reduces the escalation rate to 8.2\% and yields ${\sim}$90\% cost savings, as detailed in Appendix~\ref{app:robustness}.


\subsection{Ablation Study: Gatekeeper Efficacy and Sensitivity}
\label{sec:ablation}

\begin{figure}[t]
  \centering
  \includegraphics[width=0.99\linewidth,keepaspectratio]{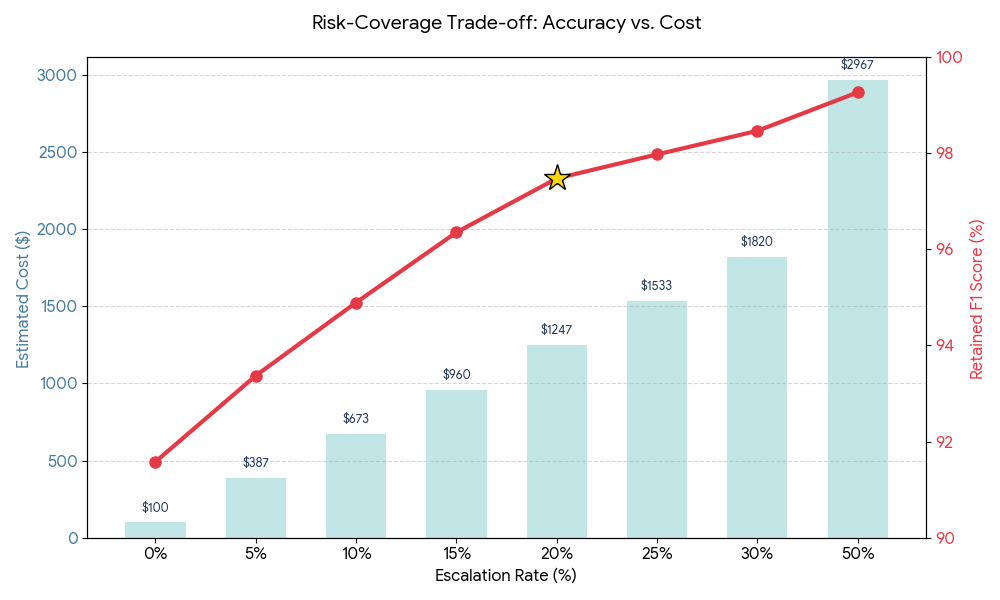}
  \caption{Risk-Coverage Trade-off. Retained F1 and operational cost as a function of escalation rate.}
  \label{fig:risk_tradeoff}
  \vspace{-4mm}
\end{figure}

We conduct a Risk-Coverage analysis to determine how effectively the predictive entropy identifies error-prone samples and calibrate the optimal uncertainty threshold ($\tau$).

\paragraph{Sensitivity and Optimal Trade-off.}
As visualized in Figure~\ref{fig:risk_tradeoff}, the efficacy of ACE hinges on the Gatekeeper's ability to concentrate errors within the high-uncertainty bracket. The Retained F1 curve measures the Gatekeeper's standalone performance exclusively on the subset of claims not routed to the LLM. The steep ascent of this curve confirms that predictive entropy is a robust proxy for error probability. Filtering out just the top 15\% of uncertain samples results in a dramatic performance surge in the retained subset. Notably, the system achieves a sweet spot at an escalation rate of 20\% (marked by the star), where the Gatekeeper maintains near-perfect reliability (97.47\% F1) on the remaining 80\% of the data. This implies that the vast majority of patent claims contain clear structures that do not require expensive reasoning. By offloading only this specific 20\% slice of ambiguous cases ($\tau \approx 0.0762$) to the Stage 2 LLM, the ACE framework maximizes resource efficiency while ensuring that the encoder-processed majority maintains legal-grade precision.


\subsection{Ablation Study: Component Analysis of the Expert LLM}
\label{sec:ablation_llm}

While the Gatekeeper ensures efficiency, the reliability of ACE is ultimately anchored in the Stage 2 Expert LLM. To isolate the contributions of our training strategies, we evaluate the impact of Supervised Fine-Tuning (SFT) and the Chain of Patent Thought (CoPT) protocol.

\paragraph{Impact of Domain-Specific SFT.}
As evidenced in Table \ref{tab:main_results}, the Llama-3-70B base model yields 67.75\% F1 in the zero-shot setting. Introducing two few-shot demonstrations (one Pass, one Fail) degrades performance further to 34.85\%, as patent claims routinely exceed 2,000 tokens and even minimal demonstrations consume a disproportionate share of the context, biasing the model toward the examples rather than the target claim. This confirms that in-context learning is ill-suited for long-document legal tasks. In sharp contrast, SFT boosts F1 to 94.63\%, a +26.88\% gain confirming that parametric knowledge internalization is essential for patent validation. Model scale is equally critical: a Llama-3-8B expert integrated into ACE Hybrid achieves only 89.3\% F1, failing to surpass the Gatekeeper alone (91.58\% F1), as the smaller model lacks sufficient reasoning capacity to correctly resolve the escalated high-entropy samples.

\paragraph{Necessity and Efficiency of the CoPT Protocol.}
Although CoPT yields a modest +0.13\% F1 increment, its primary contribution is operational rather than metric-driven: in patent prosecution, a single unverified error can incur substantial litigation costs. CoPT transforms the model into a transparent auditor by externalizing reasoning through element parsing and 35 U.S.C. statutory mapping, providing patent attorneys with actionable grounds for rejection. Furthermore, CoPT delivers a counter-intuitive efficiency gain: by imposing a strict JSON schema, it reduces inference time from 11.85s (unstructured SFT) to 6.88s, curtailing uncontrolled verbosity while maintaining legal robustness. A representative output is provided in Appendix~\ref{sec:appendix_copt_example}.

\section{Conclusion}
\label{sec:conclusion}

We proposed ACE to resolve the rigidity-resource dilemma in patent claim validation. By integrating an uncertainty-aware gatekeeper with an expert LLM executing a schema-constrained CoPT protocol, ACE decouples reasoning depth from evaluation cost, achieving the best F1 among evaluated methods (94.95\%) while reducing costs by 78\% and per-claim latency by 42\%. The entropy-based routing threshold transfers to real USPTO prosecution data without re-calibration, outperforming all baselines while reducing inference time by 60\%. Beyond patents, treating model uncertainty as a routing signal offers a generalizable blueprint for high-stakes domains requiring adaptive verification depth. 
\section*{Limitations} 

This work targets structural validity verification under the definiteness requirement of 35 U.S.C. \S112(b). The legal adequacy of patent claims is evaluated across multiple additional statutory axes, including novelty (\S102), non-obviousness (\S103), written description and enablement (\S112(a)), and patent-eligible subject matter (\S101). Each of these requirements entails a distinct reasoning structure: \S102/\S103 demand external comparison against prior art documents; \S112(a) requires inference of support relationships between the full specification and the claims; and \S101 involves abstract jurisprudential judgment under frameworks such as the Alice/Mayo two-step test. These external-reference-based and case-law-interpretive patentability determinations constitute fundamentally different reasoning regimes from the claim-internal structural consistency verification addressed herein, and each forms an independent research problem.

ACE is trained on ACE-40k, constructed via rule-based synthetic error injection. This approach enables controlled experimentation over rejection types codified in the MPEP; however, real examination environments include borderline cases in which examiner interpretive judgment is involved in \S112(b) rejections. This reflects the fact that our error taxonomy is designed to target explicit structural defects, and that the domain of interpretive indeterminacy, where inter-examiner disagreement arises, lies outside its intended scope. The residual error analysis on ACE-Real112b (Table 4) empirically corroborates this boundary: 20.3\% of undetected cases are attributable to subjective interpretation rather than explicit structural flaws.

Additionally, the current routing mechanism relies on an empirically calibrated threshold rather than a learned decision policy. Replacing this with a cost-aware routing function that jointly optimizes accuracy and computational budget remains future work. While the entropy-driven routing architecture is domain-agnostic in principle, the CoPT checklist is tailored to U.S. patent law; adapting to other jurisdictions (e.g., EPO Article 84) requires remapping the statutory checklist, which remains unexplored.

Finally, the verification scope of this work is confined to individual claims and cross-reference consistency within a single claim set. In practice, \S112(b) rejections frequently interact with term definitions in the specification, referential consistency with the drawings, and new matter issues arising from amendments during prosecution history. Such cross-document referential adequacy judgments exceed the designed input scope of the present framework, which operates exclusively over claim text. Extension toward an integrated verification system encompassing the full application document constitutes a separate research direction.

\section*{Ethical Considerations}
During the preparation of this work, the author(s) utilized generative AI to refine linguistic clarity and support the creation of certain diagrams. The author(s) carefully reviewed all outputs and maintain full responsibility for the intellectual content and originality of the final paper.

\bibliography{custom}

@manual{mpep,
  title        = {Manual of Patent Examining Procedure ({MPEP})},
  author       = {{USPTO}},
  edition      = {9th},
  year         = {2024},
  organization = {U.S. Department of Commerce},
  url          = {https://www.uspto.gov/web/offices/pac/mpep/}
}

@misc{mpep2173,
  author    = {{USPTO}},
  title     = {Manual of Patent Examining Procedure ({MPEP}),
               Section 2173: Claims Must Particularly Point Out
               and Distinctly Claim the Invention},
  year      = {2024},
  publisher = {United States Patent and Trademark Office},
  note      = {9th Edition, Revision 01.2024},
  url       = {https://www.uspto.gov/web/offices/pac/mpep/s2173.html},
  
}

@book{faber2023,
  author    = {Faber, Robert C. and Kowalski, Thomas J.},
  title     = {Faber \& Kowalski on Mechanics of Patent Claim Drafting},
  publisher = {Practising Law Institute},
  year      = {2023},
  edition   = {8th}
}

@inproceedings{zhao2019moverscore,
  title={MoverScore: Text Generation Evaluating with Contextualized Embeddings and Earth Mover Distance},
  author={Zhao, Wei and Peyrard, Maxime and Liu, Fei and Gao, Yang and Meyer, Christian M and Eger, Steffen},
  booktitle={Proceedings of the 2019 Conference on Empirical Methods in Natural Language Processing and the 9th International Joint Conference on Natural Language Processing (EMNLP-IJCNLP)},
  pages={563--578},
  year={2019}
}

@misc{uspto_bulk,
  author = {{USPTO}},
  title = {USPTO Bulk Data Storage System},
  year = {2025},
  url = {https://bulkdata.uspto.gov/},
  note = {Accessed: 2025-01-15}
}

@article{liu2023lost,
  title={Lost in the middle: How language models use long contexts},
  author={Liu, Nelson F and Lin, Kevin and Hewitt, John and Paranjape, Ashwin and Bevilacqua, Michele and Petroni, Fabio and Liang, Percy},
  journal={Transactions of the Association for Computational Linguistics},
  volume={12},
  pages={157--173},
  year={2024}
}

@article{zhang2023causal,
  title   = {Causal Reasoning and Large Language Models:
             Opening a New Frontier for Causality},
  author  = {K{\i}c{\i}man, Emre and Ness, Robert
             and Sharma, Amit and Tan, Chenhao},
  journal = {Transactions on Machine Learning Research},
  year    = {2024},
  url     = {https://openreview.net/forum?id=mqoxLkX210}
}

@inproceedings{suzgun2023hupd,
    title={The Harvard {USPTO} Patent Dataset: A Large-Scale, Well-Structured, and Multi-Purpose Corpus of Patent Applications},
    author={Suzgun, Mirac and Melas-Kyriazi, Luke and Sarkar, Suproteem K. and Kominers, Scott Duke and Shieber, Stuart M.},
    booktitle={Advances in Neural Information Processing Systems},
    volume={36},
    year={2023}
}

@inproceedings{tam2023ambiguity,
    title     = "Faithfulness Tests for Natural Language Explanations",
    author    = "Atanasova, Pepa and
                 Camburu, Oana-Maria and
                 Lioma, Christina and
                 Lukasiewicz, Thomas and
                 Simonsen, Jakob Grue and
                 Augenstein, Isabelle",
    booktitle = "Proceedings of the 61st Annual Meeting of the
                 Association for Computational Linguistics
                 (Volume 2: Short Papers)",
    year      = "2023",
    publisher = "Association for Computational Linguistics",
    pages     = "283--294",
    url       = "https://aclanthology.org/2023.acl-short.25/"
}

@inproceedings{zheng2023judging,
  title={Judging LLM-as-a-Judge with MT-Bench and Chatbot Arena},
  author={Zheng, Lianmin and Chiang, Wei-Lin and Sheng, Ying and Zhuang, Siyuan and Wu, Zhanghao and Zhuang, Yonghao and Lin, Zi and Li, Zhuohan and Li, Dacheng and Xing, Eric and others},
  booktitle={Advances in Neural Information Processing Systems},
  volume={36},
  pages={46595--46623},
  year={2023}
}

@inproceedings{ribeiro2020checklist,
    title = {Beyond Accuracy: Behavioral Testing of {NLP} Models with {CheckList}},
    author = {Ribeiro, Marco Tulio and Wu, Tongshuang and Guestrin, Carlos and Singh, Sameer},
    booktitle = {Proceedings of the 58th Annual Meeting of the Association for Computational Linguistics},
    month = jul,
    year = {2020},
    publisher = {Association for Computational Linguistics},
    url = {https://aclanthology.org/2020.acl-main.442},
    pages = {4902--4912}
    }

@inproceedings{sellam2020bleurt,
  title={BLEURT: Learning Robust Metrics for Text Generation},
  author={Sellam, Thibault and Das, Dipanjan and Parikh, Ankur P},
  booktitle={Proceedings of the 58th Annual Meeting of the Association for Computational Linguistics},
  pages={7881--7892},
  year={2020},
  url = "https://aclanthology.org/2020.acl-main.704/"
}

@inproceedings{liu2023geval,
    title     = "{G}-Eval: {NLG} Evaluation using {GPT}-4
                 with Better Human Alignment",
    author    = "Liu, Yang and
                 Iter, Dan and
                 Xu, Yichong and
                 Wang, Shuohang and
                 Xu, Ruochen and
                 Zhu, Chenguang",
    booktitle = "Proceedings of the 2023 Conference on Empirical
                 Methods in Natural Language Processing",
    year      = "2023",
    pages     = "2511--2522",
    publisher = "Association for Computational Linguistics",
    url       = "https://aclanthology.org/2023.emnlp-main.153/"
}

@inproceedings{fu2024gptscore,
    title     = "{GPTScore}: Evaluate as You Desire",
    author    = "Fu, Jinlan and
                 Ng, See-Kiong and
                 Jiang, Zhengbao and
                 Liu, Pengfei",
    booktitle = "Proceedings of the 2024 Conference of the
                 North American Chapter of the Association for
                 Computational Linguistics: Human Language
                 Technologies (Volume 1: Long Papers)",
    year      = "2024",
    pages     = "6556--6576",
    publisher = "Association for Computational Linguistics",
    url       = "https://aclanthology.org/2024.naacl-long.365/"
}

@inproceedings{min2023factscore,
    title = "{FActScore}: Fine-grained Atomic Evaluation of Factual Precision in Long-form Text Generation",
    author = "Min, Sewon and
      Krishna, Kalpesh and
      Lyu, Xinxi and
      Lewis, Mike and
      Yih, Wen-tau and
      Koh, Pang Wei and
      Iyyer, Mohit and
      Zettlemoyer, Luke and
      Hajishirzi, Hannaneh",
    booktitle = "Proceedings of the 2023 Conference on Empirical Methods in Natural Language Processing",
    year = "2023",
    pages = "12076--12100",
    publisher = "Association for Computational Linguistics",
    url = "https://aclanthology.org/2023.emnlp-main.741/"
}

@article{lee2020patentbert,
  title     = {Patent classification by fine-tuning {BERT} language model},
  author    = {Lee, Jieh-Sheng and Hsiang, Jieh},
  journal   = {World Patent Information},
  volume    = {61},
  pages     = {101965},
  year      = {2020},
  publisher = {Elsevier}
}

@inproceedings{banerjee2005meteor,
  title={METEOR: An automatic metric for MT evaluation with improved correlation with human judgments},
  author={Banerjee, Satanjeev and Lavie, Alon},
  booktitle={Proceedings of the ACL workshop on intrinsic and extrinsic evaluation measures for machine translation and/or summarization},
  pages={65--72},
  year={2005},
  url = "https://aclanthology.org/W05-0909/"

}

@inproceedings{zhang2019bertscore,
  title={BERTScore: Evaluating Text Generation with BERT},
  author={Zhang, Tianyi and Kishore, Varsha and Wu, Felix and Weinberger, Kilian Q and Artzi, Yoav},
  booktitle={International Conference on Learning Representations (ICLR)},
  year={2020},
  url = "https://openreview.net/forum?id=SkeHuCVFDr"
}

@inproceedings{papineni2002bleu,
  title={Bleu: a method for automatic evaluation of machine translation},
  author={Papineni, Kishore and Roukos, Salim and Ward, Todd and Zhu, Wei-Jing},
  booktitle={Proceedings of the 40th annual meeting of the Association for Computational Linguistics},
  pages={311--318},
  year={2002},
  url = {https://aclanthology.org/P02-1040/},
  doi = {10.3115/1073083.1073135}

}

@inproceedings{lin2004rouge,
  title={ROUGE: A Package for Automatic Evaluation of Summaries},
  author={Lin, Chin-Yew},
  booktitle={Text Summarization Branches Out},
  pages={74--81},
  year={2004},
  publisher={Association for Computational Linguistics},
  url = {https://aclanthology.org/W04-1013/}
}

@inproceedings{lee2024checkeval,
    title = "{C}heck{E}val: A Reliable {LLM}-as-a-Judge Framework for Evaluating Text Generation Using Checklists",
    author = "Lee, Yukyung and
      Kim, Joonghoon and
      Kim, Jaehee and
      Cho, Hyowon and
      Kang, Jaewook and
      Kang, Pilsung and
      Kim, Najoung",
    booktitle = "Proceedings of the 2025 Conference on Empirical Methods in Natural Language Processing",
    year = "2025",
    publisher = "Association for Computational Linguistics",
    url = "https://aclanthology.org/2025.emnlp-main.796/"
}

@inproceedings{yoo2025patentscore,
    title = "PatentScore: Multi-dimensional Evaluation of {LLM}-Generated Patent Claims",
    author = "Yoo, Yongmin and
      Xu, Qiongkai and
      Cao, Longbing",
    booktitle = "Proceedings of the 2025 Conference on Empirical Methods in Natural Language Processing",
    year = "2025",
    publisher = "Association for Computational Linguistics",
    url = "https://aclanthology.org/2025.emnlp-main.1564/",
    pages = {30727--30746},
    doi   = {10.18653/v1/2025.emnlp-main.1564}

}

@article{knappich2025pedantic,
  title     = {PEDANTIC: A Dataset for the Automatic Examination of Definiteness in Patent Claims},
  author    = {Valentin Knappich and Annemarie Friedrich and Anna Haetty and Simon Razniewski},
  journal   = {arXiv preprint arXiv:2505.21342},
  year      = {2025},
  note = {PatentSemTech@SIGIR 2025}

}

@inproceedings{schuster2022calm,
  title={Confident Adaptive Language Modeling},
  author={Schuster, Tal and Fisch, Adam and Gupta, Jai and Dehghani, Mostafa and Bahri, Dara and Tran, Vinh and Tay, Yi and Metzler, Donald},
  booktitle={Advances in Neural Information Processing Systems},
  volume={35},
  year={2022}
}

@article{chen2024frugalgpt,
  title={FrugalGPT: How to Use Large Language Models While Reducing Cost and Improving Performance},
  author={Chen, Lingjiao and Zaharia, Matei and Zou, James},
  journal={Transactions on Machine Learning Research},
  year={2024}
}

@inproceedings{varshney2022selective,
    title = "Investigating Selective Prediction Approaches Across Several Tasks in {IID}, {OOD}, and Adversarial Settings",
    author = "Varshney, Neeraj  and
      Mishra, Swaroop  and
      Baral, Chitta",
    editor = "Muresan, Smaranda  and
      Nakov, Preslav  and
      Villavicencio, Aline",
    booktitle = "Findings of the Association for Computational Linguistics: ACL 2022",
    month = may,
    year = "2022",
    address = "Dublin, Ireland",
    publisher = "Association for Computational Linguistics",
    url = "https://aclanthology.org/2022.findings-acl.158/",
    doi = "10.18653/v1/2022.findings-acl.158",
    pages = "1995--2002",
}
\bibliographystyle{acl_natbib}

\appendix
\appendix

\section{Representative Data Samples}
\label{sec:appendix_data}

Table~\ref{tab:dataset_example_appendix} presents detailed examples of the constructed dataset. Each invalid sample contains a targeted error injected based on the MPEP guidelines, such as Antecedent Basis violations or Logical conflicts. Detailed dataset statistics are provided in Table~\ref{tab:dataset} of the main text.

\begin{table}[h]
\centering
\small
\resizebox{0.95\columnwidth}{!}{
\begin{tabular}{p{0.1\columnwidth} p{0.75\columnwidth} p{0.15\columnwidth}}
\toprule
\textbf{Label} & \textbf{Claim Text} & \textbf{Error Type} \\
\midrule
Pass &
A semiconductor device comprising a substrate, a gate structure formed over the substrate, and a spacer adjacent to gate structure. &
None \\
\addlinespace
Fail &
A semiconductor device comprising \textbf{the} gate structure formed over the substrate, wherein the substrate is configured to receive the gate structure of claim~1. &
Antecedent \\
\addlinespace
Fail &
The system of claim~1, wherein the processor of claim~3 further executes the module of claim~7. &
Dependency \\
\addlinespace
Fail &
A battery including a charging unit that generates electrical energy from the discharged battery cell. &
Logical \\
\bottomrule
\end{tabular}
}
\caption{Representative samples from the ACE dataset. Invalid claims contain single targeted errors injected based on MPEP guidelines.}
\label{tab:dataset_example_appendix}
\end{table}

\section{Error Taxonomy Definitions}
\label{sec:error_taxonomy}

The error taxonomy employed in the ACE dataset is rigorously grounded in specific MPEP sections. For each category, we define the violation and provide a minimal example below:

\begin{enumerate}
    \item \textbf{Antecedent Basis (MPEP \S 2173.05(e)):} Referencing an element using a definite article without prior introduction. \\
    \textit{Ex: "The sensor..." appearing without a prior "a sensor".}
    
    \item \textbf{Dependency (MPEP \S 608.01(n)):} Invalid claim references, including forward referencing or circular dependencies. \\
    \textit{Ex: Claim 2 citing "The device of claim 5" (where claim 5 is subsequent).}
    
    \item \textbf{Logical (MPEP \S 2173.05(q)):} Internal contradictions or physically impossible functional relationships. \\
    \textit{Ex: "A transparent layer made of opaque metal".}
    
    \item \textbf{Ambiguity (MPEP \S 2173.05(b)):} Use of subjective or undefined degree terms that obscure the scope of the claim. \\
    \textit{Ex: "Heating to a \underline{substantially high} temperature".}
    
    \item \textbf{Syntax (35 U.S.C. 112):} Violations of formal grammatical structure or required patent formatting. \\
    \textit{Ex: Missing the transitional phrase "comprising" or incorrect punctuation.}
\end{enumerate}

\section{Synthetic Data Generation Details}
\label{sec:appendix_data_generation}

To construct the ACE dataset, we employed a rule-driven synthetic error injection method using Large Language Models (LLMs), specifically Qwen/Qwen3-32B and Gemma-3-27B-IT. To ensure the injected errors accurately reflect real-world patent prosecution challenges, we utilized Few-Shot Prompting, providing the models with specific examples of MPEP violations.

\subsection{Error Injection Prompts with In-Context Examples}
\label{sec:appendix_error_prompts}

The following templates demonstrate the instructions provided to the generator. Each prompt includes a definition of the error based on the Manual of Patent Examining Procedure (MPEP) and a minimal example to guide the generation process.

\paragraph{1. Antecedent Basis Error Prompt (MPEP \S 2173.05(e))}
\begin{quote}
\ttfamily
\small
\textbf{ROLE:} Senior Patent Prosecutor \\
\textbf{TASK:} Rewrite the provided claim set to introduce 5-15 \textbf{Antecedent Basis Errors}. \\
\textbf{DEFINITION:} An antecedent error occurs when a claim element is introduced with a definite article (e.g., "the lever", "said lever") without having been previously introduced with an indefinite article (e.g., "a lever"). \\
\textbf{EXAMPLE:} \\
\textit{[Original]} A device comprising \underline{a sensor}; wherein \underline{the sensor} captures data. \\
\textit{[Error]} A device comprising \underline{a processor}; wherein \underline{the sensor} captures data. ("the sensor" lacks a precedent "a sensor") \\
\textbf{REQUIREMENTS:} \\
1. Preserve claim structure and numbering. \\
2. Replace valid references with unmatched elements (e.g., "the handle" when no handle exists). \\
3. Output ONLY the revised claims. \\
\textbf{TARGET CLAIMS:} \{claim\_text\}
\end{quote}

\paragraph{2. Dependency Error Prompt (MPEP \S 608.01(n))}
\begin{quote}
\ttfamily
\small
\textbf{ROLE:} Senior Patent Prosecutor \\
\textbf{TASK:} Rewrite the claim set to introduce 5-15 \textbf{Dependency Errors}. \\
\textbf{DEFINITION:} A dependency error involves incorrect cross-referencing, such as referencing a non-existent claim or a future claim (forward referencing). \\
\textbf{EXAMPLE:} \\
\textit{[Original]} The method of \underline{claim 1}, further comprising... \\
\textit{[Error]} The method of \underline{claim 8}, further comprising... (If claim 8 does not exist or is subsequent). \\
\textbf{REQUIREMENTS:} \\
1. Modify the "claim X" references to invalid numbers. \\
2. Create circular dependencies (e.g., Claim 2 cites Claim 3, Claim 3 cites Claim 2). \\
3. Output ONLY the revised claims. \\
\textbf{TARGET CLAIMS:} \{claim\_text\}
\end{quote}

\paragraph{3. Logical Error Prompt (MPEP \S 2173.05(q))}
\begin{quote}
\ttfamily
\small
\textbf{ROLE:} Senior Patent Prosecutor \\
\textbf{TASK:} Rewrite the claim set to introduce 5-15 \textbf{Logical Errors}. \\
\textbf{DEFINITION:} Logical errors include internal contradictions, physical impossibilities, or inconsistent element properties. \\
\textbf{EXAMPLE:} \\
\textit{[Original]} A \underline{transparent} glass layer acting as a window. \\
\textit{[Error]} A \underline{transparent} glass layer made of \underline{opaque} steel. (Contradiction: transparent vs. opaque steel). \\
\textbf{REQUIREMENTS:} \\
1. Insert conflicting adjectives or impossible functional relationships. \\
2. Do not alter the grammatical structure, only the semantic logic. \\
3. Output ONLY the revised claims. \\
\textbf{TARGET CLAIMS:} \{claim\_text\}
\end{quote}

\paragraph{4. Ambiguity Error Prompt (MPEP \S 2173.05(b))}
\begin{quote}
\ttfamily
\small
\textbf{ROLE:} Senior Patent Prosecutor \\
\textbf{TASK:} Rewrite the claim set to introduce 5-15 \textbf{Ambiguity (Indefiniteness) Errors}. \\
\textbf{DEFINITION:} Claims must be definite. Errors involve subjective terms of degree without metric definitions. \\
\textbf{EXAMPLE:} \\
\textit{[Original]} Heating the water to \underline{100 degrees Celsius}. \\
\textit{[Error]} Heating the water to a \underline{substantially high temperature}. ("Substantially high" is indefinite). \\
\textbf{REQUIREMENTS:} \\
1. Replace precise numerical values with vague terms (e.g., "about", "approximately", "strong", "large"). \\
2. Output ONLY the revised claims. \\
\textbf{TARGET CLAIMS:} \{claim\_text\}
\end{quote}

\paragraph{5. Syntax Error Prompt (35 U.S.C. 112)}
\begin{quote}
\ttfamily
\small
\textbf{ROLE:} Senior Patent Prosecutor \\
\textbf{TASK:} Rewrite the claim set to introduce 5-15 \textbf{Syntax Errors}. \\
\textbf{DEFINITION:} Violations of formal claim formatting, punctuation, or missing transitional phrases. \\
\textbf{EXAMPLE:} \\
\textit{[Original]} A system \underline{comprising:} a processor\underline{;} and a memory. \\
\textit{[Error]} A system \underline{includes} a processor a memory (Missing "comprising", missing semicolons/commas). \\
\textbf{REQUIREMENTS:} \\
1. Remove or corrupt transitional phrases (comprising, consisting of). \\
2. Delete necessary punctuation (semicolons, periods). \\
3. Output ONLY the revised claims. \\
\textbf{TARGET CLAIMS:} \{claim\_text\}
\end{quote}

\section{Training and Experimental Details}
\label{sec:appendix_training}

All experiments were implemented using PyTorch on a single NVIDIA H100 80GB GPU environment. This section provides the complete hyperparameter configurations for both components of the ACE framework.

\subsection{Gatekeeper Model (Stage 1)}

Table~\ref{tab:gatekeeper_hyperparams} details the hyperparameters for the Gatekeeper. The objective was to ensure high-throughput screening while maintaining robust error detection capabilities.

\begin{table}[h]
\centering
\caption{Hyperparameters for Gatekeeper Training}
\label{tab:gatekeeper_hyperparams}
\resizebox{0.95\linewidth}{!}{%
\begin{tabular}{lcc}
\toprule
\textbf{Parameter} & \textbf{Value} & \textbf{Rationale} \\
\midrule
Base Model & PatentBERT-base & Domain-Adapted Efficiency \\
Max Sequence Length & 512 & Standard BERT Limit \\
Batch Size & 32 & Stable Gradient Estimation \\
Learning Rate & $2 \times 10^{-5}$ & Optimal for Fine-tuning \\
Epochs & 5 & Sufficient Convergence \\
Weight Decay & $0.01$ & L2 Regularization \\
\bottomrule
\end{tabular}
}
\end{table}

\subsection{Expert LLM Evaluator (Stage 2)}

For the LLM Evaluator, we employed Parameter-Efficient Fine-Tuning (PEFT) using QLoRA to adapt the model to the legal domain while minimizing memory overhead. Table~\ref{tab:llm_hyperparams} summarizes the configuration.

\begin{table}[h]
\centering
\caption{Hyperparameters for Llama-3 Evaluator Training (QLoRA)}
\label{tab:llm_hyperparams}
\resizebox{0.95\linewidth}{!}{%
\begin{tabular}{lcc}
\toprule
\textbf{Parameter} & \textbf{Value} & \textbf{Rationale} \\
\midrule
Base Model & Meta-Llama-3-70B-Instruct & Instruction-Following Capability \\
Quantization & 4-bit (NF4) & Memory Efficiency (QLoRA) \\
LoRA Rank ($r$) & 16 & Parameter Efficiency \\
LoRA Alpha ($\alpha$) & 32 & Scaling Factor ($2 \times r$) \\
LoRA Dropout & 0.05 & Regularization \\
Learning Rate & $2 \times 10^{-4}$ & Optimal for QLoRA Fine-tuning \\
Optimizer & Paged AdamW (32-bit) & Memory Optimization \\
Precision & Bfloat16 (BF16) & H100 Mixed Precision Stability \\
Batch Size & 1 (per device) & Restricted by VRAM (Context 4096) \\
Gradient Accumulation & 24 & Effective Batch Size = 24 \\
\bottomrule
\end{tabular}
}
\end{table}

\subsection{Cost Estimation Protocol}

To provide a realistic assessment of industrial scalability, we define the \textit{Estimated Cost (\$)} metric. This serves as a standardized proxy for the operational expenditure (OPEX) of deploying the model at the scale of a national patent office. The cost is calculated as:
\begin{equation}
    \text{Cost} = \frac{\text{Total Inference Time (hrs)}}{1M \text{ Claims}} \times \text{Hourly Rate}
\end{equation}
We assume a standard cloud compute rate of \$3.00 per hour for an on-demand NVIDIA H100 instance. Latency measurements include tokenization, model inference, and detokenization overheads, averaged over the entire test set.

\section{Fine-grained Performance Analysis}
\label{sec:appendix_fine_grained}

Table~\ref{tab:fine_grained_appendix} presents the detailed performance metrics of the Gatekeeper across five distinct error categories. The results reveal a clear performance dichotomy: while the model demonstrates robust detection in semantic tasks (Ambiguity, Logical), it exhibits significant limitations in structural verification tasks (Antecedent, Syntax). This structural deficiency empirically justifies the necessity of the Stage 2 LLM Evaluator for handling complex long-range dependencies.

\begin{table}[h]
\centering
\caption{Fine-grained Performance of the Gatekeeper (Test Set). The model struggles with structural errors (Antecedent, Syntax), highlighting the specific areas where the LLM Evaluator is most needed.}
\label{tab:fine_grained_appendix}
\resizebox{0.95\linewidth}{!}{%
\begin{tabular}{lccc}
\toprule
\textbf{Error Type} & \textbf{Recall} & \textbf{F1-Score} & \textbf{Nature} \\
\midrule
\multicolumn{4}{l}{\textbf{Semantic Reasoning Tasks}} \\
\midrule
Ambiguity & 90.75\% & 0.8985 & Semantic Interpretation \\
Logical & 86.45\% & 0.8645 & Inferential Reasoning \\
\midrule
\multicolumn{4}{l}{\textbf{Structural Verification Tasks}} \\
\midrule
Dependency & 92.37\% & 0.9130 & Claim Hierarchy \\
Antecedent & \textbf{82.77\%} & \textbf{0.8141} & Long-Range Reference \\
Syntax & \textbf{78.35\%} & \textbf{0.8340} & Grammatical Structure \\
\bottomrule
\end{tabular}
}
\end{table}

\section{Prompt Specifications}
\label{sec:appendix_prompts}

We provide the exact prompt templates used for our experiments. To ensure reproducibility, all prompts were formatted using the Llama-3 specific chat template structure (e.g., \texttt{<|start\_header\_id|>system...}).

\subsection{Standard SFT Prompt (Baseline)}
\label{sec:sft_prompt}

For the \textbf{SFT Only} setting, we aligned the inference prompt exactly with the training data format to maximize the effect of the fine-tuned adapter. The model is instructed to function as a "Senior Patent Attorney" and output a concise verdict without reasoning, mirroring its training objective. The exact prompt template is provided in Table~\ref{tab:sft_prompt}.

\begin{table}[h]
    \centering
    \footnotesize
    \renewcommand{\arraystretch}{1} 
    \setlength{\tabcolsep}{3pt}       
    \begin{tabular*}{\linewidth}{@{\extracolsep{\fill}} p{0.15\linewidth} | p{0.79\linewidth} @{}}
        \toprule
        \textbf{Role} & \textbf{Content} \\
        \midrule
        \textbf{System} & You are a Senior Patent Attorney. \\
        & Your task is to evaluate the validity of a patent claim based on specific error types. \\
        & If ANY error exists, output \texttt{\{"verdict": "Fail"\}}. If NO error, output \texttt{\{"verdict": "Pass"\}}. \\
        \midrule
        \textbf{User} & Claim: \textit{\{claim\_text\}} \\
        \midrule
        \textbf{Assistant} & \textit{(Model Generation Target)} \\
        \bottomrule
    \end{tabular*}
    \caption{The Standard SFT prompt template. This strictly follows the format used during the supervised fine-tuning stage to prevent distribution shift during inference.}
    \label{tab:sft_prompt}
\end{table}

\subsection{Chain of Patent Thought (CoPT) Protocol}
\label{sec:appendix_copt}

To enable the SFT model to perform deep reasoning despite being trained on short answers, we utilized a specific \textbf{System Prompt Override} strategy. The detailed prompt structure is presented in Table~\ref{tab:copt_prompt}.

\subsubsection{System Instruction Design}
The system prompt maintains the training trigger phrase ("You are a Senior Patent Attorney") to activate the fine-tuned weights. However, we explicitly injected a \texttt{[Special Instruction]} tag to override the default behavior, mandating the model to follow a step-by-step reasoning chain before concluding.

\subsubsection{In-Context Demonstrations (2-Shot)}
We employed a 2-shot In-Context Learning strategy to demonstrate the expected JSON structure.
\begin{itemize}
    \item \textbf{Shot 1 (Pass):} Demonstrates valid element parsing and statutory compliance.
    \item \textbf{Shot 2 (Fail):} Demonstrates how to identify an Antecedent Basis error.
\end{itemize}
These examples specifically teach the model to parse elements (Step 1) and check for statutory compliance (Step 2) before rendering a verdict.

\begin{table*}[t!]
\centering
\small
\renewcommand{\arraystretch}{1.2}
\setlength{\tabcolsep}{4pt} 
\begin{tabular*}{\linewidth}{@{\extracolsep{\fill}} p{0.12\linewidth} | p{0.84\linewidth} @{}}
\toprule
\textbf{Role} & \textbf{Content} \\
\midrule
\textbf{System} & You are a Senior Patent Attorney. \\
& Your task is to evaluate the validity of a patent claim based on specific error types. \\
& \\
& \textbf{[Special Instruction]:} To ensure strict adherence to 35 U.S.C. \S~112, you must explicitly document your reasoning process using the Chain of Patent Thought (CoPT) protocol before concluding. \\
& \\
& 1. \textbf{[Element Parsing]}: Identify preamble, transition, and body elements. \\
& 2. \textbf{[Statutory Compliance]}: Check for Antecedent Basis errors, Indefinite terms, and Logical Contradictions. \\
& 3. \textbf{[Verdict]}: Determine if the claim passes or fails based strictly on the above. \\
& \\
& \textbf{Output Requirement}: \\
& You must return a JSON object containing a ``reasoning'' field (for the step-by-step analysis) and a ``verdict'' field (``Pass'' or ``Fail''). \\
& Format: \texttt{\{"reasoning": "Step 1... Step 2...", "verdict": "Pass/Fail"\}} \\
\midrule
\textbf{User (Shot 1)} & Claim: ``A semiconductor device comprising a substrate, a gate structure formed over the substrate, and a spacer adjacent to the gate structure.'' \\
\midrule
\textbf{Assistant (Shot 1)} & \texttt{\{"reasoning": "Step 1 [Element Parsing]: The claim includes a substrate, a gate structure, and a spacer. Step 2 [Statutory Compliance]: 'the substrate' and 'the gate structure' have clear antecedent bases from the preamble. No indefinite terms or logical conflicts found. Step 3 [Verdict]: The claim meets all 35 U.S.C. \S~112 requirements.", "verdict": "Pass"\}} \\
\midrule
\textbf{User (Shot 2)} & Claim: ``A device comprising a processor; wherein the sensor captures data based on input.'' \\
\midrule
\textbf{Assistant (Shot 2)} & \texttt{\{"reasoning": "Step 1 [Element Parsing]: The claim introduces a processor but explicitly recites 'the sensor' in the body. Step 2 [Statutory Compliance]: The term 'the sensor' lacks a proper antecedent basis (i.e., 'a sensor' was never introduced). This is a violation of 35 U.S.C. \S~112(b). Step 3 [Verdict]: The claim is invalid due to lack of antecedent basis.", "verdict": "Fail"\}} \\
\midrule
\textbf{User (Target)} & Claim: ``\textit{\{claim\_text\}}'' \\
\bottomrule
\end{tabular*}
\caption{The full prompt template used for the Stage 2 Specialist LLM (SFT + CoPT). The \texttt{[Special Instruction]} tag and 2-shot examples are designed to override the model's training bias towards short responses, enforcing a detailed legal analysis path.}
\label{tab:copt_prompt}
\end{table*}

\paragraph{Use of AI Assistants.}
AI assistants were used only for limited writing and language-editing support, as well as preparation of submission materials. All technical content, methodology, implementation, experimental results, and final verification were performed and checked by the authors.

\section{CoPT Output Example}
\label{sec:appendix_copt_example}

Table~\ref{tab:copt_example} presents a representative output of the Chain of Patent Thought (CoPT) protocol on a real USPTO \S112(b) rejection case (Application No. 14902826). The structured four-step reasoning enables transparent auditability by explicitly identifying the defective claim elements and citing the applicable legal basis.

\begin{table*}[t]
\centering
\small
\renewcommand{\arraystretch}{1.2}
\setlength{\tabcolsep}{6pt}
\begin{tabular}{p{2.2cm} p{12.5cm}}
\toprule
\textbf{Component} & \textbf{Content} \\
\midrule
\textbf{Input Claim} \newline (Excerpt) &
\texttt{1. (canceled)} \newline
\texttt{2. The workpiece made according to the method of claim 8, further comprising an intermediate layer containing TiAlN provided between the TixSi1-xN layer and the substrate body of the workpiece.} \newline
\texttt{(...)} \newline
\texttt{5. The workpiece made according to the method of claim 8, further comprising an intermediate layer containing CryAl1-yN provided between the TixSi1-xN layer and the substrate body of the workpiece (...)} \newline
\texttt{(...)} \newline
\texttt{8. A method for coating a workpiece having a coating comprising at least one TixSi1-xN layer, wherein x \(\leq\)0.85 (...), the method comprising: using a sputtering process to produce the TixSi1-xN layer (...)} \\
\midrule
\textbf{CoPT Output} &
\texttt{\{"reasoning": "Step 1: Parsed claim elements. Step 2: Antecedent Basis error: Claim 2 refers to claim 8, but claim 8 is not yet defined. Claim 5 refers to claim 8, but claim 8 is not yet defined. Step 3: No Definiteness errors. Step 4: Fail due to Antecedent Basis errors.", "verdict": "Fail"\}} \\
\midrule
\textbf{Ground Truth} & Fail (\S112(b) rejection in USPTO Office Action) \\
\bottomrule
\end{tabular}
\caption{Example of CoPT protocol output on a real USPTO \S112(b) case. The model identifies forward-reference dependencies (claims 2 and 5 referencing the later-defined claim 8) through structured step-by-step reasoning, providing actionable grounds for the rejection verdict.}
\label{tab:copt_example}
\end{table*}

\section{Additional Baseline Experiments}
\label{sec:appendix_baselines}

To further validate the design choices of the ACE framework, we report two additional experiments: an ablation with a smaller expert model and an evaluation of a proprietary LLM.

\paragraph{Smaller Expert Model (8B Ablation).}
We fine-tuned Llama-3-8B-Instruct under the same SFT protocol used for the 70B expert and evaluated it on ACE-40k. As shown in Table~\ref{tab:8b_ablation}, the 8B model achieves 83.5\% F1 as a standalone system (\$667/M). When integrated as the Stage 2 expert within the ACE Hybrid framework, it yields 89.3\% F1 (\$233/M). Critically, this falls below the Stage 1 Gatekeeper alone (91.58\% F1), indicating that the 8B model lacks sufficient reasoning capacity to correctly resolve the high-entropy samples routed to it. Instead of complementing the Gatekeeper, the smaller expert overrides correct predictions with incorrect verdicts, empirically confirming that the escalated subset contains genuinely complex legal ambiguities requiring 70B-class reasoning depth.

\begin{table}[t]
    \centering
    \small
    \renewcommand{\arraystretch}{1.15}
    \setlength{\tabcolsep}{5pt}
    \begin{tabular}{l c c}
    \toprule
    \textbf{Configuration} & \textbf{F1 (\%)} & \textbf{Cost (\$/M)} \\
    \midrule
    Llama-3-8B SFT (Standalone) & 83.5 & \$667 \\
    ACE Hybrid w/ 8B Expert & 89.3 & \$233 \\
    \midrule
    ACE Gatekeeper (Stage 1 Only) & 91.58 & \$100 \\
    ACE Hybrid w/ 70B Expert & \textbf{94.95} & \$1,247 \\
    \bottomrule
    \end{tabular}
    \caption{Effect of expert model scale on ACE Hybrid performance (ACE-40k test set). The 8B expert degrades below the Gatekeeper alone, confirming that 70B-class reasoning is necessary for the escalated high-uncertainty subset.}
    \label{tab:8b_ablation}
\end{table}

\paragraph{Proprietary LLM Baseline (GPT-4o).}
To provide a proprietary reference point, we evaluated GPT-4o in a zero-shot setting on ACE-Real112b (204 real USPTO \S112(b) rejections). GPT-4o achieves only 18.0\% recall, substantially below ACE Hybrid (66.2\%) and even the standalone Gatekeeper (56.4\%). This result demonstrates that neither model scale nor proprietary pretraining is sufficient for reliable \S112(b) detection without domain-specific fine-tuning and structured prompting.

\section{Threshold Sensitivity Analysis}
\label{sec:appendix_threshold}

To assess the robustness of the uncertainty-based routing under varying operational budgets, we evaluate ACE on ACE-Real112b across multiple escalation ratios. The threshold calibrated on ACE-40k is applied without re-tuning; only the percentile cutoff is varied.

\begin{table}[t]
    \centering
    \small
    \renewcommand{\arraystretch}{1.1}
    \setlength{\tabcolsep}{0pt} 
    \begin{tabular*}{0.95\columnwidth}{@{\extracolsep{\fill}}cccc}
    \toprule
    \textbf{Escal Ratio} & \textbf{Stage-2 Samples} & \textbf{Recall} & \textbf{ Cost (\$/M)} \\
    \midrule
    10\% & 20 & 60.8 & \$673 \\
    20\%  & 40 & 65.7 & \$1,247 \\
    30\% & 61 & 72.1 & \$1,820 \\
    40\% & 81 & 78.9 & \$2,393 \\
    50\% & 101 & 86.3 & \$2,967 \\
    \bottomrule
    \end{tabular*}
    \caption{Threshold sensitivity on ACE-Real112b. The recall-cost trade-off is smooth and monotonic across escalation ratios.}
    \label{tab:threshold_sensitivity}
\end{table}

As shown in Table~\ref{tab:threshold_sensitivity}, increasing the escalation ratio yields consistent recall improvements with proportional cost increases. The absence of abrupt performance transitions around the default operating point (20\%) confirms that the entropy-based routing provides a stable and predictable control surface for practitioners to adjust based on their cost-recall requirements.

\section{Robustness Analysis}
\label{app:robustness}

To ensure the reliability of ACE in practical applications, we evaluated its robustness from two complementary perspectives: internal stability via cross-validation and ecological validity under industrial-scale class imbalance.

\paragraph{Internal Stability (Cross-Validation).}
To confirm that our uncertainty-based routing is not an artifact of a specific data split, we evaluated the Gatekeeper using stratified 5-fold cross-validation. The results demonstrate high stability in the uncertainty estimation, with the Area Under the Risk-Coverage Curve (AURC) showing a negligible standard deviation across folds ($\sigma < 0.002$). Crucially, the optimal escalation rate ($\gamma^*$) consistently converged around 20\% ($\pm 1.5\%$) across all folds. This confirms that the sweet spot identified in Section~\ref{sec:ablation} is an intrinsic property of the Gatekeeper's discriminative capability rather than a fluctuation dependent on data distribution.

\paragraph{Ecological Validity (Industrial Scenarios).}
The primary experiments in Section~\ref{sec:performance_analysis} utilized a balanced dataset (1:1 Pass/Fail ratio). However, real-world patent prosecution environments are characterized by a highly skewed distribution. To assess ecological validity, we conducted a simulation using a skewed test set with a 9:1 ratio (Pass:Fail), reflecting a scenario in which the majority of incoming claims are structurally well-formed. Under this setting, the Gatekeeper processes the vast majority of valid claims via the Fast Path, dropping the LLM escalation rate from 20.3\% (balanced) to 8.2\% (skewed). Consequently, the estimated operational cost decreases to ${\sim}$\$600 per million claims, achieving a ${\sim}$90\% cost reduction compared to the standalone Expert LLM. This confirms that ACE is not merely a theoretical construct but a strictly cost-optimized solution tailored for high-volume industrial pipelines.

\end{document}